\pdfoutput=1

\documentclass[11pt]{article}

\usepackage{emnlp2021}

\usepackage{times}
\usepackage{latexsym} 
\usepackage{tabularx}
\usepackage{caption}
\usepackage{subcaption}
\usepackage{float}
\usepackage{amssymb}
\usepackage{amsmath}
\usepackage{booktabs}
\usepackage{multirow}
\usepackage{tikz}
\usepackage{graphicx}
\graphicspath{images}
\usepackage{pgfplots}
\usepackage[inline]{enumitem}
\usepackage{url}

\usepackage{soul}
\usepackage{xcolor}
\usepackage[ruled,vlined]{algorithm2e}
\pgfplotsset{compat=1.17}

\definecolor{beaublue}{rgb}{0.74, 0.83, 0.9}
\definecolor{bisque}{rgb}{1.0, 0.89, 0.77}

\usepackage[T1]{fontenc}

\usepackage[utf8]{inputenc}

\usepackage{microtype}

\newcommand\blfootnote[1]{%
  \begingroup
  \renewcommand\thefootnote{}\footnote{#1}%
  \addtocounter{footnote}{-1}%
  \endgroup
}

\newcommand\Mark[1]{\textsuperscript#1}

\usepackage[most]{tcolorbox}
\tcbset{on line, 
        boxsep=4pt, left=0pt,right=0pt,top=0pt,bottom=0pt,
        colframe=white,colback=beaublue,  
        highlight math style={enhanced}
        }

\title{Graph Reasoning with Context-Aware Linearization for\\ Interpretable Fact Extraction and Verification}

\author{Neema Kotonya\Mark{{*1}}, \hfill Thomas Spooner\Mark{2}, \hfill Daniele Magazzeni\Mark{2},\and Francesca Toni\Mark{1} \\

\Mark{1}Department of Computing, Imperial College London \\ \Mark{2}J.P. Morgan AI Research \\ 
nk2418@ic.ac.uk, \{thomas.spooner,daniele.magazzeni\}@jpmorgan.com, ft@ic.ac.uk

}
  
\begin{document}
\maketitle
\begin{abstract}
\blfootnote{* Work done while the author was an intern at J.P. Morgan AI Research.}
This paper presents an end-to-end system for fact extraction and verification using textual and tabular evidence, the performance of which we demonstrate on the FEVEROUS dataset. We experiment with both a multi-task learning paradigm to jointly train a graph attention network for both the task
of evidence extraction and veracity prediction, as well as a single objective graph model for solely learning veracity prediction and separate evidence extraction. In both instances, we employ a framework for per-cell linearization of tabular evidence, thus allowing us to treat evidence from tables as sequences. The templates we employ for linearizing tables capture the context as well as the content of table data. We furthermore provide a case study to show the interpretability our approach. Our best performing system achieves a FEVEROUS score of 0.23 and 53\% label accuracy  on the blind test data.\footnote{This system was not submitted to the shared task competition, but instead to 
\href{https://eval.ai/web/challenges/challenge-page/1091/leaderboard/2806}{the after
competition leader board} under the name \textbf{CARE} (\textbf{C}ontext \textbf{A}ware \textbf{RE}asoner).}

\end{abstract}

\section{Introduction}

Fact checking has become an increasingly important tool to combat misinformation. Indeed the study of automated fact checking in NLP \cite{vlachos-riedel-2014-fact}, in particular, has yielded a number of valuable insights in recent times. These include
task formulations such as
matching for discovering already fact-checked claims \cite{shaar-etal-2020-known}, identifying neural fake news \cite{zellers2020defending}, fact verification in scientific \cite{wadden-etal-2020-fact} and public health \cite{kotonya-toni-2020-explainable-automated} domains, and end-to-end fact verification \cite{thorne-etal-2018-fever}, which is the subject of the FEVEROUS benchmark dataset \cite{aly2021feverous}.

A majority of automated fact checking studies only consider text as evidence for verifying claims. Recently, there have been a number of works which look at fact-checking
with 
structured and semi-structured data, mainly in the form of tables and knowledge bases \cite{2019TabFactA} ---
but fact-checking from both structured and unstructured data has been largely unexplored. 
Given the sophistication in the presentation of fake news, it is important to develop fact checking tools for assessing evidence from a wide array of evidence sources in order to reach a more accurate verdict regarding the veracity of 
claims.


In this work, we propose a graph-based representation that supports both textual and tabular evidence, thus addressing some of the key limitations of past architectures.
This approach allows us to capture relations between evidence items as well as claim-evidence
pairs, borrowing from the argumentation and argument mining literature \citep{argmining-2020-argument,vecchi-etal-2021-towards}, as well as argument modeling for fact verification \cite{alhindi-etal-2018-evidence}.

We experiment with two formulations for graph learning. For the first, we employ a multi-task learning paradigm to jointly train  a  graph  attention  network \cite{velivckovic2017graph}  for  both  the task of evidence extraction --- which we model as a node  selection  task --- and  a graph-level  veracity prediction task. In the second, we explicitly separate the verification and extraction tasks, where standard semantic search is used for evidence extraction, and veracity prediction is treated as a graph-level classification problem. 

For veracity prediction we predict a label for each claim, one of \textsc{Supports}, \textsc{Refutes}, or \textsc{Not-Enough-Info} (NEI), which is conditioned on all relevant evidence, hence the intuition to frame veracity prediction as a graph-level prediction task.
In both formulations, we employ context-aware table linearization templates to produce per-cell sequence representations of tabular evidence and thus construct evidence reasoning graphs where nodes have heterogeneous evidence types (i.e., representing sentences and tables on the same evidence reasoning graph).

\paragraph{Contributions.}
The three main contributions of the paper are summarized below:
\begin{enumerate}
    \item Provide \textbf{insightful empirical analysis} of the new FEVEROUS benchmark dataset.
    \item Propose a novel framework for interpretable fact extraction using templates to derive \textbf{context-aware per-cell linearizations}.
    \item Present a \textbf{graph reasoning model} for fact verification that supports both structured and unstructured evidence data.
\end{enumerate}

Both the joint model and separately trained models exhibit a significant improvement over the FEVEROUS baseline, as well as significant improvements for label accuracy and evidence recall. Our separated approach to fact extraction and verification achieves a FEVEROUS score of 0.23 and label accuracy of 53\% on the blind test data.

\section{Related Work}\label{sec:related-work}
\paragraph{Graph Reasoning for Fact  Verification.} Several works explore graph neural networks (GNN) for fact extraction and verification, both for fine-grained evidence modelling \cite{liu-etal-2020-fine,zhong-etal-2020-reasoning} and evidence aggregation for veracity prediction \cite{zhou-etal-2019-gear}. Furthermore, graph learning has also been leveraged to build fake news detection models which learn from evidence from different contexts; e.g., user-based and content-based data \cite{liu-etal-2020-fine,lu-li-2020-gcan}. There are also non-neural approaches to fake news detection with graphs \cite{AhmadiLPS19,Kotonya-toni-2019-gradual}. However, to the best of our knowledge, this work is the first to employ a graph structure to jointly reason over both text and tabular evidence data in both single task learning (STL) and multi-task learning (MTL) settings. 
\paragraph{Table Linearization.} A number of approaches have been adopted in NLP for table linearization. For example, \citet{gupta-etal-2020-infotabs} study natural language inference in the context of table linearizations, in particular they are interested to see if language models can infer entailment relations from table linearizations. The linearization approach employed by \citet{Schlichtkrull-etal-2021-joint} 
is also used for automated fact verification. However, 
they linearize tables 
row- and column-wise, whereas we focus on cell
s
as evidence items in the FEVEROUS dataset are annotated at table-cell level. 

\section{Data Analysis}\label{sec:data}

Further to the FEVEROUS dataset statistics discussed by the task description paper \cite{aly2021feverous}, we perform our own data exploration. We present insights from our data analysis of the FEVEROUS dataset, which we use to inform system design choices.

\paragraph{Table types.} Wikipedia tables can be categorized into one of two classes: infoboxes and general tables. Infoboxes are fixed format tables which typically appear in the top right-hand corner of a Wikipedia article.  General tables can convey a wider breadth of information (e.g., election results, sports match scores, the chronology of an event)
and typically  have more complex structures (e.g., multiple headers).
List items can also be considered as a special subclass of tables, where the number of items is analogous to the number of columns
and the nests of the list signify table rows.

\paragraph{Evidence types.} The first observation we make is that, similar to the FEVER dataset \cite{thorne-etal-2018-fever}, a sizeable portion of the training instances rely on evidence items which are extracted from the first few sentences of a Wikipedia article. The most common evidence items are the first and second sentences in a Wikipedia article, which appear in 36\% and 18\% of evidence sets, respectively. The four most frequent evidence cells all come from the first table, with 49\% of first tables listed as evidence in the train and dev data being infoboxes. Further, the vast majority of cell evidence items are non-header cells, but these only account  for approximately 5.1\% of tabular evidence in the train and dev datasets. A summary of these findings is provided in Table \ref{tab:table-type} for the most common evidence types in the training data.


\begin{table}[ht]
    \centering
    \begin{tabular}{lr}
    \toprule
    \textbf{ Evidence type} & \textbf{\% Evidence sets} \\
    \midrule 
     List items & 1.6\% \\
     Sentences & 67.7\% \\
     All tables  & 58.2\%\\
      \hspace*{3mm} Infoboxes  & 26.5\%\\
       \hspace*{3mm} General tables  & 33.9\%\\
    \bottomrule
    \end{tabular}
    \caption{Prevalence of evidence types in the training data by number of evidence sets in which they appear.}
    \label{tab:table-type}
\end{table}

\paragraph{Evidence item co-occurrences.} We investigate the most common evidence pairs, both in individual evidence sets and also in the union of all evidence sets relating to a claim. The most common evidence pair in the training data is (\textsc{sentence\_0}, \textsc{sentence\_1}), which accounts for 3.2\% of evidence co-occurrences. The most common sentence-table cell co-occurrence is (\textsc{cell\_0\_2\_1}, \textsc{sentence\_0}). The most common table cell pair is (\textsc{cell\_0\_2\_0}, \textsc{cell\_0\_2\_1}). All of the ten most common co-occurrences either contain one of the first four sentences in an article or evidence from one of the first two tables.  


\paragraph{\textsc{NEI} label.} Lastly, we choose to explore instances of the NEI class. We sample 100 instances of NEI claims from the training data and note their qualitative attributes. We
pay particular attention to this label as it is the least represented in the data. Unlike the FEVER score, the FEVEROUS metric requires the correct evidence, as well as the label, to be supplied for an NEI instance for credit to awarded. Our analysis is summarized in Table \ref{tab:NEI-categories}. We categorize mutations, using the FEVEROUS  annotation scheme, as one of three types: entity substitution
, including more facts than available in the provided evidence (i.e., including additional propositions), and paraphrasing or generalizing.  We use \emph{Other} to categorize claims with a mutation not captured by one of these three categories.

\begin{table}[ht]
    \centering
    \begin{tabular}{p{5cm}r}
    \toprule
      \textbf{Mutation Type}   & \textbf{\% Sample} \\
      \midrule
       Entity Substitution  & 21\%  \\
       More facts than in evidence  &  42\% \\ 
       Paraphrasing or generalizing & 36\% \\
      Other & 1\% \\
       
    \bottomrule
    \end{tabular}
    \caption{We sample 100 NEI instances and categorize them according to the type of lexical mutation which results in the claim being unverifiable.}
    \label{tab:NEI-categories}
\end{table}

We note that a number of NEI examples are mutations of \textsc{Supports} or \textsc{Refutes} examples. For example the claim in Table \ref{tab:NEI} is a mutation of a \textsc{Supports} instance where entity substitution (humans $\rightarrow$ reptiles) has been used to make the first clause unverifiable, hence changing the label to NEI.

\begin{table}[ht]
    \begin{tabular}{p{7.2cm}}
    \hline

    \begin{center}
    \textbf{Claim} \\
    \sethlcolor{beaublue}  

    \hl{Nucleoporin 153, a protein which in \textbf{reptiles} is encoded by the NUP153 gene},\\
    \sethlcolor{bisque}
    \hl{is an essential component of the basket of nuclear pore complexes (NPCs) in vertebrates, and required for the anchoring of NPCs.}\\[1em]
    \end{center}
    
    \\[1em]\hline

    \begin{center}
    
    \textbf{Evidence} \\
    \sethlcolor{beaublue}  \hl{
    Nucleoporin 153 (Nup153) is a protein which in \textbf{humans} is encoded by the NUP153 gene.}\\
    
    \sethlcolor{bisque}
    \hl{
    It is an essential component of the basket of nuclear pore complexes (NPCs) in vertebrates, and required for the anchoring of NPCs.}
    
    \end{center}
    \\[1em]\hline

    \end{tabular}
    \caption{NEI example where the evidence is highlighted according to the part of the claim to which it refers. The text in \textbf{bold} is the substitution which resulted in the label changing from \textsc{Supports} to \text{NEI}.}
    \label{tab:NEI}
\end{table}

\section{Methods}

\begin{figure*}
    \centering
    \includegraphics[width=\textwidth]{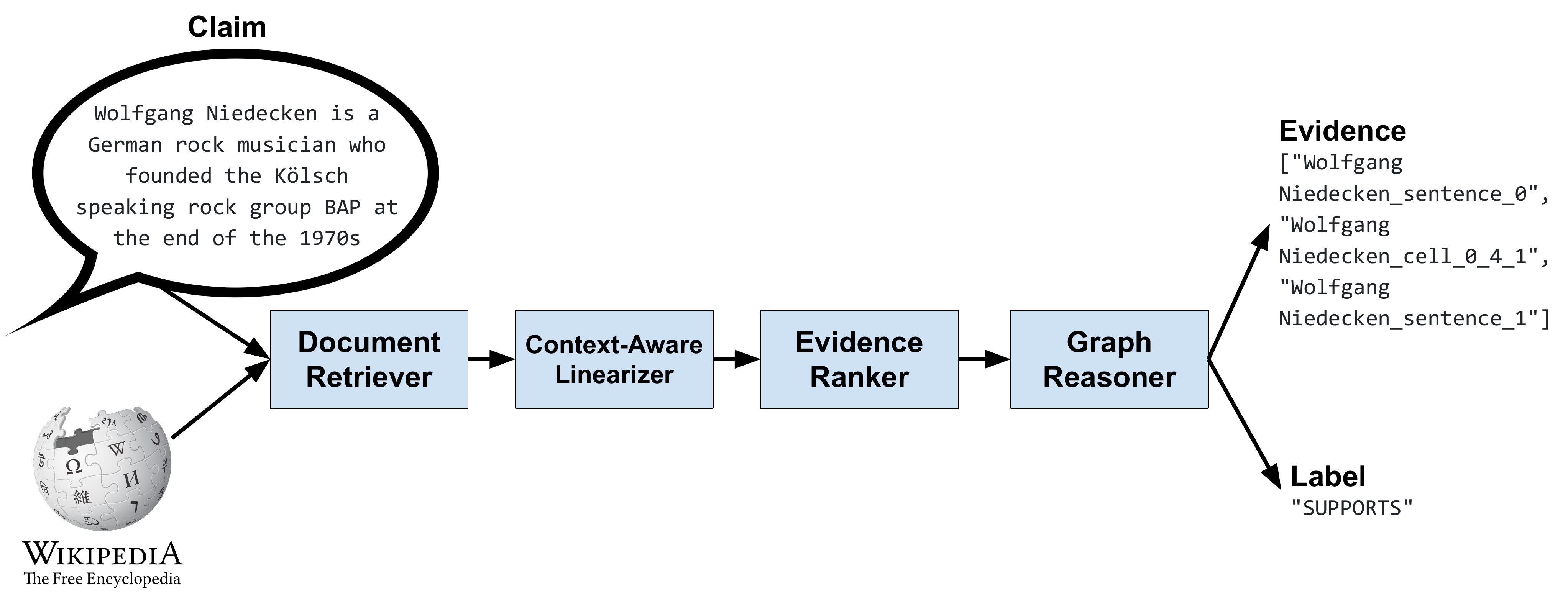}
    \caption{Our fact verification pipeline. We employ two graph reasoning approaches: STL where the evidence extraction and modelled separately, and MTL where further evidence filtering is performed jointly with veracity prediction by the Graph Reasoner.}
    \label{fig:system}
\end{figure*}

Our proposed method for fact verification is an end-to-end system comprising 
three modules:
\begin{enumerate}[label=(\arabic*)]
    \item A robust document retrieval procedure (see Section~\ref{sec:doc_retreival}).
    \item An evidence graph construction and intermediate evidence filtering process (see Section~\ref{sec:graph_construction})
    .
    \item A joint veracity label prediction and evidence selection layer that reasons over the evidence graph (see Section~\ref{sec:graph_reasoning}).
\end{enumerate}
An illustration of the complete pipeline is provided in Figure~\ref{fig:system}, and details of each processing stage are
provided in the following sections.

\subsection{Document Retrieval}\label{sec:doc_retreival}

For document retrieval, we employ an entity linking and API search approach
similar to that of \citet{hanselowski-etal-2018-ukp}. The {WikiMedia API}\footnote{\url{https://www.mediawiki.org/wiki/API}} is used to query Wikipedia for articles related to the claim, using named entities and noun phrases from the claim as search terms. These retrieved Wikipedia page titles form our candidate document set.
Named entities that are not retrieved by the API are then extracted from the claim
as a handful of these identify pages which are present in the Wikipedia dump (e.g., \textbf{/wiki/Lars\_Hjorth} is present in the provided Wikipedia evidence dump, but is not returned by the WikiMedia API). In the same vein, we discard titles which are returned by the API, but are not in the Wikipedia dump. TF-IDF and cosine similarity are employed to score and rerank the retrieved Wikipedia articles with respect to their similarity to the claim.

As in the approach of \citet{hanselowski-etal-2018-ukp}, the seven highest ranked pages are chosen at test time.
For completeness, we also experiment with approaches to document retrieval which select pages based on a threshold score \citep{nie2019revealing}. Ultimately, we find these methods yield lower precision. 

\subsection{Evidence Reasoning Graph}\label{sec:graph_construction}

Similar to other fact verification systems \cite{augenstein-etal-2019-multifc,hidey-etal-2020-deseption}, we jointly train our model for both the evidence selection and veracity prediction tasks. In contrast to these approaches, however, we employ a graph reasoning module for the joint learning of the two tasks.
We choose this approach to exploit the permutation invariance of evidence with respect to a claim, as there is no canonical ordering of evidence. Our graph formulation differs from previous graph-based fact verification systems in that 
we construct a \emph{heterogeneous graph to model both tabular and sequence evidence data}.

In the following sections we will describe two specific approaches that are taken for the fact verification task:
\begin{enumerate*}[label=(\arabic*)]
    \item where we condition the graph model to learn both node-level, fine-grained evidence selection and graph-level veracity label prediction simultaneously, and
    \item where we only learn graph-level veracity prediction.
\end{enumerate*}

\sethlcolor{beaublue} 
\begin{table*}[ht]
    \centering
    \scalebox{0.9}{
    \begin{tabular}{cp{5cm}p{7.5cm}}
    \toprule
    \textbf{Evidence Type} &  \textbf{Linearization} & \textbf{Example from FEVEROUS dataset}
     \\\midrule 
    \textbf{Infoboxes} & & \\
    \cmidrule{1-1}
     Headers  &  \texttt{TABLE} \textbf{has} \tcbox{\texttt{CELL\_I\_J}} \newline [in \texttt{SUBHEADER}] & Brewster Productions has \tcbox{Genres}. \newline [\textbf{/wiki/Brewster\_Productions}] \\[0.2cm]
    Non-headers    &   \texttt{CELL\_I\_0} \textbf{of} \texttt{TABLE} \newline [in \texttt{SUBHEADER}] \textbf{is}  \tcbox{\texttt{CELL\_I\_J}} & Current ranking of Barbora Krejčíková in Singles is \tcbox{No. 65 (16 November 2020)}. [\textbf{/wiki/Barbora\_Krejcikova}]\\
    \midrule
     \textbf{General tables} &  &\\
     \cmidrule{1-1}
    Headers  &  \texttt{TABLE} \textbf{has} \tcbox{\texttt{CELL\_I\_J}} \newline [in \texttt{SUBHEADER}] & The 1964 United States Senate election in Maine has \tcbox{Party}.\newline [\textbf{/wiki/1908\_Clemson\_Tigers\_football\_team}]\\[0.2cm]
    Non-headers & \texttt{TABLE/PAGE} \textbf{has} \newline \texttt{SUBHEADER\_0} \texttt{CELL\_I\_0}\newline \textbf{in}  \texttt{SUBHEADER\_J} \newline\textbf{of} \tcbox{\texttt{CELL\_I\_J}} & 2014 Ladies European Tour has Rank 9 in Player of \tcbox{Florentyna Parker}.\newline [\textbf{/wiki/2014\_Ladies\_European\_Tour}]\\
    \midrule
    \textbf{List items} & & \\ 
    \cmidrule{1-1}
    Without subheaders  & \texttt{TITLE} \textbf{includes} \tcbox{\texttt{ITEM\_I\_J}}  & Site includes \tcbox{Location, a point or an area on the} \tcbox{Earth's surface or elsewhere.} \newline[\textbf{/wiki/Site}] \\[0.2cm]
    With subheaders  & \texttt{SUBHEADERS} \textbf{for} \texttt{TITLE} \newline \textbf{includes} \tcbox{\texttt{ITEM\_I\_J}} & The Player Honours for Park Sang-in includes \tcbox{K-League Best XI: 1985} \newline [\textbf{/wiki/Park\_Sang-in}] \\
    \bottomrule
    \end{tabular}}
    \caption{Templates for encoding tabular evidence. \texttt{CELL\_I\_0}, \texttt{SUBHEADER\_0}, \texttt{SUBHEADER\_J}, \texttt{SUBHEADERS}, \texttt{TABLE}, \texttt{TITLE} and \texttt{PAGE} are all context elements. The content of the evidence item is  \tcbox{highlighted}. In each case \texttt{ITEM\_I\_J} denotes list item content and \texttt{CELL\_I\_J}  denotes table cell content.}
    \label{tab:linearization}
\end{table*}

\paragraph{Linearizing Tabular Data.}

We linearize both table and list evidence data and generate from these linearizations a contextualized sequence representation which captures information about each cell as well as its surrounding page elements. 
This is accomplished using templates that distinguish explicitly between infoboxes and general tables.
For the latter, we engineer the templates to handle two particular complexities that are present only in general tables:
\begin{enumerate*}[label=(\arabic*)]
    \item nested headers, and
    \item table cells which span multiple rows and multiple columns (see Figure \ref{fig:complex-tables}).
\end{enumerate*}
Furthermore, we also employ templates for producing context-rich representations of item lists (see Table \ref{tab:linearization} for more details).

\sethlcolor{pink} 
\begin{figure}[ht]
    \centering
    \scalebox{0.79}{
    \begin{tabular}{|c|c|c|c|c|}
    \hline
    \multirow{2}{*}{\tcbox{Club}} & \multirow{2}{*}{\tcbox{Season}} & \multicolumn{3}{|c|}{\tcbox{League}} \\ \cline{3-5} 
    &  & Division & Apps & Goals \\ \hline
    Santa Cruz   & 2019  & Série C & 7    & 1     \\ \hline
    \multirow{3}{*}{\tcbox{Athletico Paranaense}} & 2020 & \multirow{2}{*}{\tcbox{Serie A}} & 0  & 0  \\ \cline{2-2}\cline{4-5}
   & 2021  &  & 0    & 0     \\ \cline{2-5}
   & \multicolumn{2}{|c|}{\tcbox{Total}}  & 0    & 0  \\ \hline
   Guarani (loan)  & 2020 & Série B  & 5 & 0 \\ \hline
    \end{tabular}}
    \caption{Example of a complex general table taken from \textbf{/wiki/Elias\_Carioca}. This table contains both multi-row cells and multi-column cells, some of which are headers. They are shown \tcbox{highlighted}.}

    \label{fig:complex-tables}
\end{figure}

\paragraph{Graph Structure.} We construct a fully connected graph $G = (V, E)$, where each node $n_i \in V$ represents a claim-evidence pair, similar to previous evidence graphs for automated fact checking \cite{zhao2019transformer,zhou-etal-2019-gear}.
Self-loops are also included in $G$ for each node in order to improve evidence reasoning, so the set of edges for the graph is $ E = \{(n_i, n_j) \text{ }\vert\text{ } n_i, n_j \in V \}$.

At test time, we take
the Wikipedia pages output by the document retrieval module, segment each Wikipedia page into its constituent page items (i.e., sentences, table cells, table captions and list items), and refer to these as evidence items.
These evidence items are then filtered. Using an ensemble of pre-trained S-BERT sentence embeddings \cite{reimers-gurevych-2019-sentence},
we perform semantic search with the claim as our query. Cosine similarity is then used to rank the evidence items. For the joint and single training approaches, we select a different number of evidence nodes; in particular, a larger graph is used with the former.
For training, we select nodes to occupy the graph according to the following rule-set:
\begin{enumerate}[label=(\arabic*)]
    \item If gold evidence, include as a node.
    \item For claims that require a single evidence item, include the top four candidates returned using our semantic search approach as nodes.
    \item For claims with more than one gold evidence item, retrieve the same number of candidates as gold items.
\end{enumerate}
The union of these sets form the collection of nodes, $V$, that occupy the evidence graph $G$.


\paragraph{Node Representations.}

For the initial node representations, similar to \citet{liu-etal-2020-fine} and \citet{zhao2019transformer}, we represent evidence nodes with the claim to which they refer as context.
The claim is concatenated with a constructed context-rich evidence sequence $e_i$. When constructing the sequences, $e_i$, we consider the unstructured evidence items (i.e, sentences and table captions) and the structured table and list items separately.

For sentences and table captions the evidence sequence is generated by concatenating the evidence item with the page title which serves as context. For table cells and list items we perform a per cell linearization, where this linearization forms the evidence sequence for table and list item evidence items (see Table \ref{tab:linearization} for the templates used). For each evidence item, we feed this claim-evidence sequence pair to a RoBERTa encoder \cite{liu2019roberta}, and  each node $\mathbf{n}_i \in V$ in an evidence graph has the pooled output of the last hidden state of the [CLS] token, $\mathbf{h}_i^0$ as its initial state:

\begin{equation}
\mathbf{n}_i =  \mathbf{h}_i^0 = \text{RoBERTa}_{\text{CLS}}(c, e_i).
\label{eq:rep1}
\end{equation}



\subsection{Evidence Selection and Veracity Prediction}\label{sec:graph_reasoning}

\paragraph{Training graphs.} We train two graph networks, one for joint veracity prediction and evidence extraction, and the second solely for the veracity prediction task. 

\paragraph{Oversampling NEI Instances.}
As discussed in Section~\ref{sec:data}, the FEVEROUS dataset suffers from a significant class imbalance with respect to the NEI instances.
Similar to the baseline approach, we employ 
techniques for generating new NEI instances in order to
address this issue.
Concretely, we use two data augmentation strategies in order to increase the number of NEI at train time:
\begin{enumerate*}[label=(\arabic*)]
    \item evidence set reduction, and
    \item claim mutation.
\end{enumerate*}
For the first case, we randomly sample \textsc{Supports} and \textsc{Refutes} instances
and drop evidence. Given the distribution of entity substituted and non-entity substituted mutations --- as discovered in our data analysis (see Section \ref{sec:data}) --- we make the choice to include in the training data: 15,000 constructed NEI examples made using the first approach, and 5,946 NEI examples constructed using the second. This means that a total of 92,237 NEI examples were used for model training.

\paragraph{STL: Separate Verification and Extraction.}

For the first model, we perform the tasks of fact extraction and verification of evidence selection and veracity prediction separately. We make use of an ensemble semantic search method for extracting top evidence items for claims. We employ S-BERT\footnote{We use the `msmarco-distilbert-base-v4' and `paraphrase-mpnet-base-v2'' pretrained models.} to encode the claim and the evidence items separately. We then compute cosine similarity for the claim evidence pair. 

The 25 highest ranking tabular evidence items were chosen, and the top-scoring 5 sentences (and captions) for each claim were selected as the nodes of our evidence reasoning graph at test time. This is the evidence limit stated by the FEVEROUS metric. 

When constructing the evidence graph at test time, we choose to exclude header cells and list items evidence types as nodes as they account for a very small portion of evidence items (see Section \ref{sec:data}), and experimentation shows that the evidence extraction model has a bias to favour these evidence elements over sentences. We use two GAT layers in our graph reasoning model, with: a hidden layer size of 128, embeddings size of 1024, and a global attention layer for node aggregation. The logits generated by the model are fed directly to a categorical cross entropy loss function, and the veracity label output probability distribution \text{\textbf{p}$_i$}, for each evidence graph $G_i \in \mathcal{G} $, is computed using the relation
\begin{equation}
 \mathbf{p}_i = \mathrm{softmax}(\text{MLP}(\mathbf{W} \mathbf{o}_i  + \mathbf{b})),
\label{eq:veracity-prob}
\end{equation}
where
\begin{equation}
\mathbf{o}_i = \sum_{n_i \in V}^{n_i} \mathrm{softmax} \left(
h_{\mathrm{gate}} ( \mathbf{x}_n ) \right) \odot
h_{\mathbf{\Theta}} ( \mathbf{x}_n ).
\label{eq:global-attention}
\end{equation}


\paragraph{MTL: Joint Verification and Extraction.}
We also experiment with a joint training or multi-task learning (MTL) approach in order to explore if simultaneously learning the veracity label and evidence items can lead to improvements in the label accuracy metric and also evidence prediction recall and precision. For this approach, we construct larger evidence graphs at test time, including the thirty-five highest ranked evidence items according to the S-BERT evidence extraction module. The intention is for the graph network to learn a binary classification for each claim-evidence pair in the network. 

For the multi-task learning model, we increase the dimensions of our graph network by feeding our initial input graphs to two separate GAT components (in order to increase the model's capacity for 
learning the more complex multi-task objective), the outputs of which, $\mathbf{h}_a$ and $\mathbf{h}_b$, are concatenated to form representation $\mathbf{h}$ over which we compute global attention, 
where the combined representation takes the form:\footnote{We denote the concatenation of vectors $\mathbf{x}$ and $\mathbf{y}$, by $[\mathbf{x}; \mathbf{y}]$.}
\begin{equation}
    \mathbf{h} = [ \mathbf{h}_a ; \mathbf{h}_b ].
\end{equation}
The binary cross entropy loss is then used for the node-level evidence selection task, and, as with the separated model, we use categorical cross entropy to compute the graph-level veracity prediction, as shown in   (\ref{eq:veracity-prob}) and (\ref{eq:global-attention}). 
The resulting joint graph neural network is then trained with the linear-additive objective
\begin{equation}
    \mathcal{L}_{\text{joint}} = \lambda \mathcal{L}_{\text{evidence}} + \mathcal{L}_{\text{label}},
\end{equation}
taking the form of a Lagrangian with multiplier $\lambda \geq 0$, where
\begin{equation}
    \mathcal{L}_{\text{evidence}} = \mathrm{sigmoid}(\text{MLP}(\mathbf{W}_i \mathbf{h}  + \mathbf{b})).
\end{equation}
As with the previous approach, we feed the model logits to our loss functions and use an Adam optimizer to train the network, and set $\lambda = 0.5$.

\subsection{Hyper-parameter Settings}

For all models, we make use of a  \textsc{RoBERTA-large} model which is pre-trained on a number of NLI datasets including \textsc{NLI-FEVER} \citep{nie-etal-2020-adversarial}.  We use a maximum sequence length of 512 for encoding all claim-evidence concatenated pairs. We experiment with the following learning rates [\underline{1e-5}, 5e-5, 1e-4], ultimately choosing the learning rate underlined. Training was performed using batch size of 64. We train the single objective model for 20k steps,
choosing the weights with the minimum veracity prediction label loss, and train the joint model for 20k
steps, taking the model with highest recall for evidence extraction. The Adam optimizer is used in training for both approaches.

\section{Results}

We report the results of the entire
fact extraction and verification 
pipeline, as well as the evaluation of the pipeline's performance for
intermediate stages of the fact verification system, e.g., document retrieval and evidence selection.

\paragraph{Document retrieval.} Our method for DR shows significant improvement on the TF-IDF+DrQA approach used by the baseline. In particular we find that our document retrieval module sees gains from querying the Wikipedia dump for pages related to entities which are not retrieved by the WikiMedia API. However, we do note that our approach struggles to retrieve Wikipedia pages in cases relating to specific events which can only be inferred through reasoning over the claim. 

For example, consider the following claim from the development dataset: \textit{``2014 Sky Blue FC season number 18 Lindsi Cutshall (born October 18, 1990) played the FW position.''}. In this case, the document selection process returns \textit{``Sky Blue FC''}, \textit{``Lindsi Cutshall''}, and \textit{``2015 Sky Blue FC season''}, but does not return the gold evidence page \textit{``2014 Sky Blue FC season''} which is required for verification of the claim.

We report recall@$k$ for $k =\{3,5,7\}$ where $k$ is the number of Wikipedia page documents retrieved by the module. Our approach shows significant improvements over the baseline (see Table \ref{tab:document_retrieval}).

\begin{table}[ht]
    \centering
    \begin{tabular}{lccc}
    \toprule
    \textbf{Method} & \textbf{Rec@3} & \textbf{Rec@5} & \textbf{Rec@7}\\
    \midrule
    Baseline &  0.58    &  0.69 &  -- \\
    Ours  & 0.65  & 0.73 & \textbf{0.80}\\
    \hline
    \end{tabular}
    \caption{Document retrieval results measured by Recall@$k$, where $k$ is the number of documents retrieved. Results reported for the dev set.}
    \label{tab:document_retrieval}
\end{table}

\paragraph{Evidence selection and veracity prediction.} For evidence selection and veracity prediction, we observe that the approach trained for the single objective of veracity prediction marginally outperforms the jointly trained module (see Table \ref{tab:evidence_coverage}). We hypothesize that the difficulty of learning to select the correct evidence nodes along with predicting veracity might be the cause of this. It is possible that performance of the joint model could be improved with better evidence representation or through the use of a different graph structure, e.g., by incorporating edge attributes. 

\begin{table}[H]
    \centering
    \begin{tabular}{lcc}
    \toprule
    \textbf{Method} &  \textbf{Recall}  & \textbf{LA} \\
    \midrule
    Baseline                 &      29.51 & 53.22 \\
    STL    &   \textbf{37.20}&
    \textbf{
    62.89}\\
    MTL  & 36.25    & 62.21\\
    \bottomrule
    \end{tabular}
    \caption{System performance of the  dev set for evidence recall and label accuracy.}
    \label{tab:evidence_coverage}
\end{table}

Finally, we submitted our blind test results for STL, which is our best performing method, to the after-competition FEVEROUS leaderboard. Our system outperforms the baseline significantly on both the FEVEROUS metric and also label accuracy as reported in Table \ref{tab:final_results}. Furthermore, our results on the blind test data show almost no degradation from development to test set with respect to the evidence recall which remains at 37\%. So the cause of our reduced FEVEROUS score between the development and test data is mainly due to a decrease in label accuracy from 63\% on the development data to 53\% for the test data. We are confident that this could be improved with better label accuracy for the NEI class.

\begin{table}[ht]
    \centering
    
    \begin{tabular}{ccccc}
    \hline
     \multirow{2}{*}{\textbf{Method}}    & \multicolumn{2}{c}{\textbf{Dev}}  & \multicolumn{2}{c}{\textbf{Test}} \\
     \cmidrule{2-5}
       & \textbf{LA} & \textbf{FS} & \textbf{LA} & \textbf{FS} \\
       \hline
      Baseline & 53.22 & 19.28 & 47.60 & 17.70 \\ 
      Ours & 
      \textbf{62.81} & \textbf{25.71} & \textbf{53.12} & \textbf{22.51} \\
      \bottomrule
    \end{tabular}
    \caption{Results for label accuracy (\textbf{LA}) and FEVEROUS score (\textbf{FS}) for the full pipeline on both the development and blind test datasets.}
    \label{tab:final_results}
\end{table}



\subsection{Case Study and System Interpretability}

We present an example of a claim from the development dataset, which requires both tabular and textual evidence to be verified. We show how it is labelled by our pipeline (see Table \ref{tab:case_study}). For this example, our evidence selection module correctly identifies all three evidence items required to fact-check the claim. Furthermore, two of the three evidence items receive the highest relevance scores from our evidence selection module. Of the irrelevant evidence items retrieved for this claim, eleven out of twenty-two come from an unrelated Wikipedia page (``Scomadi Turismo Leggera"). The correct label of \textsc{Supports} is also predicted for this instance. 

In order to explore the interpretability system predictions, for this same instance, we analyse the node attention weights for the first GAT layer, they are shown in parenthesis for each predicted evidence item in Table \ref{tab:case_study}. We can see that  the two evidence nodes with the highest values both correspond to items in the gold evidence set. However the third gold evidence item, \textsc{Scomadi\_sentence\_15}, has a much lower weight than a number of items which are not in the gold evidence set.

\begin{table}[ht]
    \centering
    \begin{tabular}{p{7.1cm}}
    \toprule
    \textbf{Claim} 
    ``In 2019, Scomadi, a private limited company with limited liability, was bought by a British owner which changed Scomadi's management structure."\\
    \midrule
    \textbf{Evidence} \\
    Scomadi\_cell\_0\_0\_1,\newline Scomadi\_sentence\_14, Scomadi\_sentence\_15.\\
    
    \textbf{Predicted Evidence}\\
    (1) \tcbox{Scomadi\_cell\_0\_0\_1}\hfill (\textit{0.1794}),\newline
    (2) \tcbox{ Scomadi\_sentence\_14}\hfill (\textit{0.1203}),\newline 
    (3) Scomadi\_table\_caption\_0\hfill (\textit{0.0871}), \newline 
    (4) Scomadi\_cell\_0\_3\_1 \hfill (\textit{0.0685}), \newline
    (5) Scomadi\_cell\_0\_7\_1\hfill (\textit{0.0561}), \newline
    (6) Scomadi\_cell\_0\_2\_1\hfill (\textit{0.0472}) \newline
    (7) Scomadi\_cell\_0\_8\_1\hfill (\textit{0.0405}) \newline
    (8) \tcbox{Scomadi\_sentence\_15}\hfill (\textit{0.0360}), \newline
    (9) Scomadi\_sentence\_11\hfill (\textit{0.0324}), \newline
    (10) Scomadi\_sentence\_0\hfill (\textit{0.0292}), \newline
    (11) Scomadi\_cell\_0\_6\_1\hfill (\textit{0.0266}), \newline
    (12) Scomadi\_cell\_0\_5\_1\hfill (\textit{0.0243}), \newline
    (13) Scomadi\_cell\_0\_1\_1\hfill (\textit{0.0224}), \newline
    (14) Scomadi\_cell\_0\_4\_1\hfill (\textit{0.0208}).\\
    \midrule
    \textbf{Label} \hfill \textsc{Supports}\\
    \textbf{Predicted Label} \hfill \textsc{Supports}\\
    \hline
    \end{tabular}
    \caption{Example claim from the development dataset which requires extracting both tabular and textual evidence in order for it to be verified. For brevity we only show the top fourteen (out of twenty-five) extracted evidence items, correctly predicted evidence is \tcbox{highlighted}.}
    \label{tab:case_study}
\end{table}


\section{Conclusion and Future Work}

In this work, we have demonstrated two novel approaches for fact extraction and verification that support both structured and unstructured evidence. These architectures were motivated by literature in argumentation, and also by the empirical analysis presented in Section~\ref{sec:data}. Our results show significant improvement over the shared task baseline for both the joint and separated models, with the latter generating a marginal improvement on the FEVEROUS metric compared with the former. Overall, we conclude that the use of graph-based reasoning in fact verification systems could hold great promise for future lines of work.

We hypothesize that exploring varied task formulations could potentially yield strong improvements in model performance, for example: constructing reasoning graphs on an evidence set level, or using the FEVER dataset to augment the NEI claims used during training, or further fine-tuning sentence embeddings on the FEVEROUS dataset. Furthermore, we believe further insights could be gained by evaluating our table linearization approach on other datasets related to fact verification over tabular data. In addition to this, we hope to conduct further experiments with our graph based approach using structured and unstructured evidence independently, to further investigate which aspect of our approach led to the improvement on the FEVEROUS score.

Incorporating prior knowledge or constraints into the training procedure would also be an interesting direction. 
Finally, we 
believe that our graph-based approach lends itself well to the extraction of veracity prediction \emph{explanations} \cite{kotonya-toni-2020-explainable}, obtained from evidence extracted from our underpinning graphs as justifications for claims. The ability to provide evidence for a claim, and to justify this, would better enable the integration of these techniques in practical systems.

\paragraph{Disclaimer}
This paper was prepared for informational purposes by the Artificial
Intelligence Research group of JPMorgan Chase \& Co and its affiliates (``J.P.\
Morgan''), and is not a product of the Research Department of J.P.\ Morgan.
J.P.\ Morgan makes no representation and warranty whatsoever and disclaims all liability, for the completeness, accuracy or reliability of the information
contained herein. This document is not intended as investment research or
investment advice, or a recommendation, offer or solicitation for the purchase
or sale of any security, financial instrument, financial product or service, or
to be used in any way for evaluating the merits of participating in any
transaction, and shall not constitute a solicitation under any jurisdiction or to any person, if such solicitation under such jurisdiction or to such person would be unlawful. \copyright{} 2021 JPMorgan Chase \& Co. All rights reserved.

\bibliography{custom,anthology}

\begin{thebibliography}{29}
\expandafter\ifx\csname natexlab\endcsname\relax\def\natexlab#1{#1}\fi

\bibitem[{Ahmadi et~al.(2019)Ahmadi, Lee, Papotti, and Saeed}]{AhmadiLPS19}
Naser Ahmadi, Joohyung Lee, Paolo Papotti, and Mohammed Saeed. 2019.
\newblock \href
  {https://truthandtrustonline.com/wp-content/uploads/2019/09/paper\_15.pdf}
  {Explainable fact checking with probabilistic answer set programming}.
\newblock In \emph{Proceedings of the 2019 Truth and Trust Online Conference
  {(TTO} 2019), London, UK, October 4-5, 2019}.

\bibitem[{Alhindi et~al.(2018)Alhindi, Petridis, and
  Muresan}]{alhindi-etal-2018-evidence}
Tariq Alhindi, Savvas Petridis, and Smaranda Muresan. 2018.
\newblock \href {https://doi.org/10.18653/v1/W18-5513} {Where is your evidence:
  Improving fact-checking by justification modeling}.
\newblock In \emph{Proceedings of the First Workshop on Fact Extraction and
  {VER}ification ({FEVER})}, pages 85--90, Brussels, Belgium. Association for
  Computational Linguistics.

\bibitem[{Aly et~al.(2021)Aly, Guo, Schlichtkrull, Thorne, Vlachos,
  Christodoulopoulos, Cocarascu, and Mittal}]{aly2021feverous}
Rami Aly, Zhijiang Guo, Michael Schlichtkrull, James Thorne, Andreas Vlachos,
  Christos Christodoulopoulos, Oana Cocarascu, and Arpit Mittal. 2021.
\newblock \href {http://arxiv.org/abs/2106.05707} {Feverous: Fact extraction
  and verification over unstructured and structured information}.

\bibitem[{Augenstein et~al.(2019)Augenstein, Lioma, Wang, Chaves~Lima, Hansen,
  Hansen, and Simonsen}]{augenstein-etal-2019-multifc}
Isabelle Augenstein, Christina Lioma, Dongsheng Wang, Lucas Chaves~Lima, Casper
  Hansen, Christian Hansen, and Jakob~Grue Simonsen. 2019.
\newblock \href {https://doi.org/10.18653/v1/D19-1475} {{M}ulti{FC}: A
  real-world multi-domain dataset for evidence-based fact checking of claims}.
\newblock In \emph{Proceedings of the 2019 Conference on Empirical Methods in
  Natural Language Processing and the 9th International Joint Conference on
  Natural Language Processing (EMNLP-IJCNLP)}, pages 4685--4697, Hong Kong,
  China. Association for Computational Linguistics.

\bibitem[{Cabrio and Villata(2020)}]{argmining-2020-argument}
Elena Cabrio and Serena Villata, editors. 2020.
\newblock \href {https://aclanthology.org/2020.argmining-1.0}
  {\emph{Proceedings of the 7th Workshop on Argument Mining}}. Association for
  Computational Linguistics, Online.

\bibitem[{Chen et~al.(2020)Chen, Wang, Jianshu~Chen, Wang, Li, Zhou, and
  Wang}]{2019TabFactA}
Wenhu Chen, Hongmin Wang, Yunkai~Zhang Jianshu~Chen, Hong Wang, Shiyang Li,
  Xiyou Zhou, and William~Yang Wang. 2020.
\newblock Tabfact : A large-scale dataset for table-based fact verification.
\newblock In \emph{International Conference on Learning Representations
  (ICLR)}, Addis Ababa, Ethiopia.

\bibitem[{Gupta et~al.(2020)Gupta, Mehta, Nokhiz, and
  Srikumar}]{gupta-etal-2020-infotabs}
Vivek Gupta, Maitrey Mehta, Pegah Nokhiz, and Vivek Srikumar. 2020.
\newblock \href {https://doi.org/10.18653/v1/2020.acl-main.210} {{INFOTABS}:
  Inference on tables as semi-structured data}.
\newblock In \emph{Proceedings of the 58th Annual Meeting of the Association
  for Computational Linguistics}, pages 2309--2324, Online. Association for
  Computational Linguistics.

\bibitem[{Hanselowski et~al.(2018)Hanselowski, Zhang, Li, Sorokin, Schiller,
  Schulz, and Gurevych}]{hanselowski-etal-2018-ukp}
Andreas Hanselowski, Hao Zhang, Zile Li, Daniil Sorokin, Benjamin Schiller,
  Claudia Schulz, and Iryna Gurevych. 2018.
\newblock \href {https://doi.org/10.18653/v1/W18-5516} {{UKP}-athene:
  Multi-sentence textual entailment for claim verification}.
\newblock In \emph{Proceedings of the First Workshop on Fact Extraction and
  {VER}ification ({FEVER})}, pages 103--108, Brussels, Belgium. Association for
  Computational Linguistics.

\bibitem[{Hidey et~al.(2020)Hidey, Chakrabarty, Alhindi, Varia, Krstovski,
  Diab, and Muresan}]{hidey-etal-2020-deseption}
Christopher Hidey, Tuhin Chakrabarty, Tariq Alhindi, Siddharth Varia, Kriste
  Krstovski, Mona Diab, and Smaranda Muresan. 2020.
\newblock \href {https://doi.org/10.18653/v1/2020.acl-main.761}
  {{D}e{S}e{P}tion: Dual sequence prediction and adversarial examples for
  improved fact-checking}.
\newblock In \emph{Proceedings of the 58th Annual Meeting of the Association
  for Computational Linguistics}, pages 8593--8606, Online. Association for
  Computational Linguistics.

\bibitem[{Kotonya and Toni(2019)}]{Kotonya-toni-2019-gradual}
Neema Kotonya and Francesca Toni. 2019.
\newblock \href {https://doi.org/10.18653/v1/W19-4518} {Gradual argumentation
  evaluation for stance aggregation in automated fake news detection}.
\newblock In \emph{Proceedings of the 6th Workshop on Argument Mining}, pages
  156--166, Florence, Italy. Association for Computational Linguistics.

\bibitem[{Kotonya and Toni(2020{\natexlab{a}})}]{kotonya-toni-2020-explainable}
Neema Kotonya and Francesca Toni. 2020{\natexlab{a}}.
\newblock \href {https://doi.org/10.18653/v1/2020.coling-main.474} {Explainable
  automated fact-checking: A survey}.
\newblock In \emph{Proceedings of the 28th International Conference on
  Computational Linguistics}, pages 5430--5443, Barcelona, Spain (Online).
  International Committee on Computational Linguistics.

\bibitem[{Kotonya and
  Toni(2020{\natexlab{b}})}]{kotonya-toni-2020-explainable-automated}
Neema Kotonya and Francesca Toni. 2020{\natexlab{b}}.
\newblock \href {https://doi.org/10.18653/v1/2020.emnlp-main.623} {Explainable
  automated fact-checking for public health claims}.
\newblock In \emph{Proceedings of the 2020 Conference on Empirical Methods in
  Natural Language Processing (EMNLP)}, pages 7740--7754, Online. Association
  for Computational Linguistics.

\bibitem[{Liu et~al.(2019)Liu, Ott, Goyal, Du, Joshi, Chen, Levy, Lewis,
  Zettlemoyer, and Stoyanov}]{liu2019roberta}
Yinhan Liu, Myle Ott, Naman Goyal, Jingfei Du, Mandar Joshi, Danqi Chen, Omer
  Levy, Mike Lewis, Luke Zettlemoyer, and Veselin Stoyanov. 2019.
\newblock \href {http://arxiv.org/abs/1907.11692} {Roberta: A robustly
  optimized bert pretraining approach}.

\bibitem[{Liu et~al.(2020)Liu, Xiong, Sun, and Liu}]{liu-etal-2020-fine}
Zhenghao Liu, Chenyan Xiong, Maosong Sun, and Zhiyuan Liu. 2020.
\newblock \href {https://doi.org/10.18653/v1/2020.acl-main.655} {Fine-grained
  fact verification with kernel graph attention network}.
\newblock In \emph{Proceedings of the 58th Annual Meeting of the Association
  for Computational Linguistics}, pages 7342--7351, Online. Association for
  Computational Linguistics.

\bibitem[{Lu and Li(2020)}]{lu-li-2020-gcan}
Yi-Ju Lu and Cheng-Te Li. 2020.
\newblock \href {https://doi.org/10.18653/v1/2020.acl-main.48} {{GCAN}:
  Graph-aware co-attention networks for explainable fake news detection on
  social media}.
\newblock In \emph{Proceedings of the 58th Annual Meeting of the Association
  for Computational Linguistics}, pages 505--514, Online. Association for
  Computational Linguistics.

\bibitem[{Nie et~al.(2019)Nie, Wang, and Bansal}]{nie2019revealing}
Yixin Nie, Songhe Wang, and Mohit Bansal. 2019.
\newblock \href {http://arxiv.org/abs/1909.08041} {Revealing the importance of
  semantic retrieval for machine reading at scale}.

\bibitem[{Nie et~al.(2020)Nie, Williams, Dinan, Bansal, Weston, and
  Kiela}]{nie-etal-2020-adversarial}
Yixin Nie, Adina Williams, Emily Dinan, Mohit Bansal, Jason Weston, and Douwe
  Kiela. 2020.
\newblock \href {https://doi.org/10.18653/v1/2020.acl-main.441} {Adversarial
  {NLI}: A new benchmark for natural language understanding}.
\newblock In \emph{Proceedings of the 58th Annual Meeting of the Association
  for Computational Linguistics}, pages 4885--4901, Online. Association for
  Computational Linguistics.

\bibitem[{Reimers and Gurevych(2019)}]{reimers-gurevych-2019-sentence}
Nils Reimers and Iryna Gurevych. 2019.
\newblock \href {https://doi.org/10.18653/v1/D19-1410} {Sentence-{BERT}:
  Sentence embeddings using {S}iamese {BERT}-networks}.
\newblock In \emph{Proceedings of the 2019 Conference on Empirical Methods in
  Natural Language Processing and the 9th International Joint Conference on
  Natural Language Processing (EMNLP-IJCNLP)}, pages 3982--3992, Hong Kong,
  China. Association for Computational Linguistics.

\bibitem[{Schlichtkrull et~al.(2021)Schlichtkrull, Karpukhin, Oguz, Lewis, Yih,
  and Riedel}]{Schlichtkrull-etal-2021-joint}
Michael~Sejr Schlichtkrull, Vladimir Karpukhin, Barlas Oguz, Mike Lewis,
  Wen-tau Yih, and Sebastian Riedel. 2021.
\newblock \href {https://doi.org/10.18653/v1/2021.acl-long.529} {Joint
  verification and reranking for open fact checking over tables}.
\newblock In \emph{Proceedings of the 59th Annual Meeting of the Association
  for Computational Linguistics and the 11th International Joint Conference on
  Natural Language Processing (Volume 1: Long Papers)}, pages 6787--6799,
  Online. Association for Computational Linguistics.

\bibitem[{Shaar et~al.(2020)Shaar, Babulkov, Da~San~Martino, and
  Nakov}]{shaar-etal-2020-known}
Shaden Shaar, Nikolay Babulkov, Giovanni Da~San~Martino, and Preslav Nakov.
  2020.
\newblock \href {https://doi.org/10.18653/v1/2020.acl-main.332} {That is a
  known lie: Detecting previously fact-checked claims}.
\newblock In \emph{Proceedings of the 58th Annual Meeting of the Association
  for Computational Linguistics}, pages 3607--3618, Online. Association for
  Computational Linguistics.

\bibitem[{Thorne et~al.(2018)Thorne, Vlachos, Christodoulopoulos, and
  Mittal}]{thorne-etal-2018-fever}
James Thorne, Andreas Vlachos, Christos Christodoulopoulos, and Arpit Mittal.
  2018.
\newblock \href {https://doi.org/10.18653/v1/N18-1074} {{FEVER}: a large-scale
  dataset for fact extraction and {VER}ification}.
\newblock In \emph{Proceedings of the 2018 Conference of the North {A}merican
  Chapter of the Association for Computational Linguistics: Human Language
  Technologies, Volume 1 (Long Papers)}, pages 809--819, New Orleans,
  Louisiana. Association for Computational Linguistics.

\bibitem[{Vecchi et~al.(2021)Vecchi, Falk, Jundi, and
  Lapesa}]{vecchi-etal-2021-towards}
Eva~Maria Vecchi, Neele Falk, Iman Jundi, and Gabriella Lapesa. 2021.
\newblock \href {https://doi.org/10.18653/v1/2021.acl-long.107} {Towards
  argument mining for social good: A survey}.
\newblock In \emph{Proceedings of the 59th Annual Meeting of the Association
  for Computational Linguistics and the 11th International Joint Conference on
  Natural Language Processing (Volume 1: Long Papers)}, pages 1338--1352,
  Online. Association for Computational Linguistics.

\bibitem[{Velickovic et~al.(2018)Velickovic, Cucurull, Casanova, Romero,
  Li{\`{o}}, and Bengio}]{velivckovic2017graph}
Petar Velickovic, Guillem Cucurull, Arantxa Casanova, Adriana Romero, Pietro
  Li{\`{o}}, and Yoshua Bengio. 2018.
\newblock \href {https://openreview.net/forum?id=rJXMpikCZ} {Graph attention
  networks}.
\newblock In \emph{6th International Conference on Learning Representations,
  {ICLR} 2018, Vancouver, BC, Canada, April 30 - May 3, 2018, Conference Track
  Proceedings}. OpenReview.net.

\bibitem[{Vlachos and Riedel(2014)}]{vlachos-riedel-2014-fact}
Andreas Vlachos and Sebastian Riedel. 2014.
\newblock \href {https://doi.org/10.3115/v1/W14-2508} {Fact checking: Task
  definition and dataset construction}.
\newblock In \emph{Proceedings of the {ACL} 2014 Workshop on Language
  Technologies and Computational Social Science}, pages 18--22, Baltimore, MD,
  USA. Association for Computational Linguistics.

\bibitem[{Wadden et~al.(2020)Wadden, Lin, Lo, Wang, van Zuylen, Cohan, and
  Hajishirzi}]{wadden-etal-2020-fact}
David Wadden, Shanchuan Lin, Kyle Lo, Lucy~Lu Wang, Madeleine van Zuylen, Arman
  Cohan, and Hannaneh Hajishirzi. 2020.
\newblock \href {https://doi.org/10.18653/v1/2020.emnlp-main.609} {Fact or
  fiction: Verifying scientific claims}.
\newblock In \emph{Proceedings of the 2020 Conference on Empirical Methods in
  Natural Language Processing (EMNLP)}, pages 7534--7550, Online. Association
  for Computational Linguistics.

\bibitem[{Zellers et~al.(2020)Zellers, Holtzman, Rashkin, Bisk, Farhadi,
  Roesner, and Choi}]{zellers2020defending}
Rowan Zellers, Ari Holtzman, Hannah Rashkin, Yonatan Bisk, Ali Farhadi,
  Franziska Roesner, and Yejin Choi. 2020.
\newblock \href {http://arxiv.org/abs/1905.12616} {Defending against neural
  fake news}.

\bibitem[{Zhao et~al.(2020)Zhao, Xiong, Rosset, Song, Bennett, and
  Tiwary}]{zhao2019transformer}
Chen Zhao, Chenyan Xiong, Corby Rosset, Xia Song, Paul Bennett, and Saurabh
  Tiwary. 2020.
\newblock \href {https://openreview.net/forum?id=r1eIiCNYwS} {Transformer-xh:
  Multi-evidence reasoning with extra hop attention}.
\newblock In \emph{International Conference on Learning Representations}.

\bibitem[{Zhong et~al.(2020)Zhong, Xu, Tang, Xu, Duan, Zhou, Wang, and
  Yin}]{zhong-etal-2020-reasoning}
Wanjun Zhong, Jingjing Xu, Duyu Tang, Zenan Xu, Nan Duan, Ming Zhou, Jiahai
  Wang, and Jian Yin. 2020.
\newblock \href {https://doi.org/10.18653/v1/2020.acl-main.549} {Reasoning over
  semantic-level graph for fact checking}.
\newblock In \emph{Proceedings of the 58th Annual Meeting of the Association
  for Computational Linguistics}, pages 6170--6180, Online. Association for
  Computational Linguistics.

\bibitem[{Zhou et~al.(2019)Zhou, Han, Yang, Liu, Wang, Li, and
  Sun}]{zhou-etal-2019-gear}
Jie Zhou, Xu~Han, Cheng Yang, Zhiyuan Liu, Lifeng Wang, Changcheng Li, and
  Maosong Sun. 2019.
\newblock \href {https://doi.org/10.18653/v1/P19-1085} {{GEAR}: Graph-based
  evidence aggregating and reasoning for fact verification}.
\newblock In \emph{Proceedings of the 57th Annual Meeting of the Association
  for Computational Linguistics}, pages 892--901, Florence, Italy. Association
  for Computational Linguistics.

\end{thebibliography}
\bibliographystyle{acl_natbib}

\end{document}